\documentclass[11pt]{article}

\usepackage[margin=1in]{geometry}
\usepackage[T1]{fontenc}
\usepackage{lmodern}
\usepackage{microtype}
\usepackage{amsmath,amssymb}
\usepackage{booktabs,array,colortbl,multirow}
\usepackage{xcolor}
\usepackage{graphicx}
\usepackage{caption}
\usepackage{tikz}
\usepackage{pgfplots}
\usepackage[numbers,sort&compress]{natbib}
\usepackage[colorlinks=true,linkcolor=blue!50!black,citecolor=blue!50!black,urlcolor=blue!50!black]{hyperref}

\usetikzlibrary{arrows.meta,positioning,calc}
\pgfplotsset{compat=1.17}
\definecolor{best}{gray}{0.88}

\title{A System~1 Decision Model for Medical Multiple-Choice Questions\\Built on Frozen Language Models}
\author{Pavan Kumar Mantha \\ Antz.ai}
\date{14 November 2026}

\begin{document}
\maketitle

\begin{abstract}
A System~1 model answers a multiple-choice question in a single pass. A frozen language model turns the question and each option into vectors, two small trained heads place them in a shared space, and the closest option wins. The design has been shown on agent decisions, but we are not aware of a study of how far it goes on exam questions that need medical knowledge, where the right option is often no more similar to the question than the wrong ones. We rebuilt the recipe from scratch for MedMCQA and trained only two projection heads and one temperature on top of frozen Qwen3.5-0.8B, Granite-4.2-3B and Qwen3-8B, using a contrastive loss with in-batch and hard negatives and a mask for repeated answers. With plain last-token vectors, both small encoders answer about one question in three correctly (33.1\% and 33.0\% on 3,842 held-out questions, where chance is 25.0\%). Writing each text inside a one-word-summary prompt and lowering the learning rate of the heads raised this to 41.7\% and 42.5\% for the two small encoders and to 46.0\% $\pm$ 0.2 (three seeds) for Qwen3-8B. Shuffling the questions against the options returns accuracy to chance, so the heads do use the question. Flan-PaLM and Med-PaLM~2 report 57.6\% and 72.3\% on MedMCQA under a different protocol, so ours is a cheap, fast baseline and not a competitive system. We also say what our evidence cannot show: validation gains were about 1.7 to 2 times the test gains, and we could not separate the effect of reading several encoder layers from the other changes.
\end{abstract}

\section{Literature Survey}
\label{sec:related}

Our work sits where three lines of research meet: models that choose among candidates by comparing vectors, the use of decoder-only language models as frozen feature extractors, and question answering on medical exams. We take them in that order and then say what is left open.

\subsection{Decision models that score candidates by similarity}

Kahneman's split between fast, automatic judgement and slow, deliberate reasoning \citep{kahneman2011} is often borrowed to describe models that commit to an answer without writing out a chain of thought. Contrastive Language Models, or CLMs \citep{kwok2026clm}, make the borrowing concrete. A state encoder and an action encoder are trained with a contrastive loss so that a state lies close to the action that was actually taken. At run time the model scores every candidate action against the state and returns the best one, so the cost of a decision is one pass over the state plus one pass over each candidate, and candidate vectors can be cached. A public rebuild of the 8B model, written for a workshop, puts two small heads on a frozen Qwen3-8B, maps the last-token hidden state of each side to 512 dimensions, and scores an option by its cosine similarity times a learned scale \citep{vizuara2026}. That rebuild is trained on agent decisions and document-grounded questions, and it credits the general idea to a closed-source model called Jev. We are not aware of an evaluation of this design on medical exam questions, which is the gap we start from.

\subsection{Contrastive objectives and the choice of negatives}

The loss behind these models descends from the noise-contrastive objective used in contrastive predictive coding \citep{oord2018cpc}. Two-tower training with a symmetric in-batch loss became standard for image and text pairs in CLIP \citep{radford2021clip}. For question answering, dense passage retrieval trains a question encoder and a passage encoder so that the right passage outranks the other passages in the batch and hard negatives retrieved for each question \citep{karpukhin2020dpr}, and Sentence-BERT applies siamese training to sentence similarity \citep{reimers2019sbert}. How negatives are chosen matters, and negatives that sit close to the anchor are more informative than random ones \citep{robinson2021hardneg}.

These systems usually draw negatives from a large pool where accidental duplicates are rare. Exam answers behave differently. Short phrases such as the name of a drug or an organ recur across thousands of questions (Section~\ref{sec:data}), so an in-batch ``negative'' is often another correct answer for a similar question. We handle this with an explicit mask on repeated answers instead of a larger batch.

\subsection{Decoder language models as frozen encoders}

A decoder-only model is trained to predict the next token, and the hidden state of its last token is not trained to summarise a sentence. Three groups of work address this. The first changes the input and not the model: PromptEOL asks the model to compress a sentence into one word, with a template of the form \texttt{This sentence: "[text]"} \texttt{means in one word:"}, and reads the hidden state at the end of the prompt, which gives usable sentence embeddings without any training \citep{jiang2024prompteol}. The second trains the decoder as an embedding model, either with synthetic contrastive data as in E5-Mistral \citep{wang2024e5} or by enabling bidirectional attention and following it with short unsupervised training as in LLM2Vec \citep{behnamghader2024llm2vec}. The third asks which layer to read. Across 32 embedding tasks, intermediate layers often beat the final layer \citep{skean2025layer}, in line with the older practice of probing intermediate layers with linear classifiers \citep{alain2016probes}.

We keep the encoder frozen and untouched, as CLM does, and change only how its states are read out: the prompt, the depth, and the head that follows.

\subsection{Medical multiple-choice question answering}

MedMCQA collects more than 190,000 questions from Indian medical entrance exams across 21 subjects \citep{pal2022medmcqa}, and MedQA collects questions from medical exams \citep{jin2020medqa}. The strongest published results come from large generative models. Flan-PaLM 540B reached 57.6\% on the MedMCQA development set in its preprint \citep{singhal2022arxiv}, and Med-PaLM~2 reports 72.3\% in its preprint \citep{singhal2023arxiv}; the journal versions are \citet{singhal2023nature} and \citet{singhal2025mpalm2}. A smaller model can also be used by showing it all options in the prompt and scoring the answer letter, a setting studied by \citet{robinson2023mcq}. Our setting differs on purpose: the question and each option are encoded separately, one pass each, and nothing is generated.

\subsection{Open question and contributions}

The literature gives us a decision model that scores candidates by similarity, several recipes for reading a decoder's states, and a hard exam benchmark. It does not tell us how far the first goes on the third when the second is done cheaply, with no encoder training. This paper measures that. We contribute a from-scratch pipeline for medical multiple-choice questions that trains only two heads and a temperature and masks repeated answers; a controlled study, selected on validation data only, of how the frozen vectors are produced and how the heads are trained; and an evaluation over three encoders and three seeds, with a shuffled-question control, simple baselines, and an account of where our evidence is weak.

\section{Dataset}
\label{sec:data}

\subsection{Source}

We start from MedMCQA \citep{pal2022medmcqa}. The public release on Hugging Face lists 193,155 questions in total: 182,822 for training, 4,183 for validation and 6,150 for testing, covering 21 subjects and about 2,400 healthcare topics, under the Apache-2.0 licence. The official test split ships without an answer key, so it cannot be scored, and the labelled validation split is the only labelled held-out data. We use that split, after the processing below, as our test set, and we carve our own validation set out of the training data. To avoid confusion, we call the official labelled split the \emph{development set} and our held-out evaluation questions the \emph{test set}.

\subsection{Processing}

Each question becomes a row with the question text, the correct answer, the wrong options (``negatives''), a subject, an answer identifier and a flag for negated wording. Five steps produce the final splits.

First, we drop every question in which any option only makes sense next to the others, such as ``all of the above'', ``both A and B'' or a bare combination of letters, since such an option cannot be embedded on its own. We match these with regular expressions. Second, the correct answer is the option marked correct, and the negatives are the remaining options with duplicates and empty strings removed; a question left with no negative is dropped. Third, the answer identifier is the CRC-32 hash of the answer after lower-casing and replacing every run of non-alphanumeric characters with a space, so two rows share an identifier exactly when their answers are the same text. Fourth, a question is flagged as negated when it contains \emph{except}, \emph{not}, \emph{false}, \emph{incorrect} or \emph{least} as a whole word.

Fifth, we build the splits. The test set is the official development set after the first three steps, with questions de-duplicated on their normalised text (3,842 of 4,183 questions remain). Training questions that match a test question after normalisation are removed, and training questions are de-duplicated too. From what is left, 4,000 questions drawn at random (seed 42) form the validation set and 134,416 form the training set. Overall, 138,416 of the 182,822 training questions (75.7\%) survive. The processed dataset is public \citep{mantha2026dataset}.

\subsection{Statistics}

Table~\ref{tab:data} gives the measured statistics of the three splits. Almost every question has four options: 99.6\% in training and validation, and 99.5\% in the test set, where the other 20 questions have three. Expected chance accuracy is therefore 25.03\% in training and 25.04\% in the test set rather than exactly 25\%.

\begin{table}[t]
\centering
\caption{Statistics of the processed dataset. ``Negated'' is the share of questions containing \emph{except}, \emph{not}, \emph{false}, \emph{incorrect} or \emph{least}.}
\label{tab:data}
\small
\setlength{\tabcolsep}{4pt}
\begin{tabular}{lrrrrrr}
\toprule
Split & Questions & Four options (\%) & Dental (\%) & Negated (\%) & Words / question & Words / answer \\
\midrule
Train      & 134,416 & 99.6 & 5.1  & 21.4 & 13.6 & 2.7 \\
Validation & 4,000   & 99.6 & 4.1  & 20.4 & 13.8 & 2.7 \\
Test       & 3,842   & 99.5 & 29.4 & 19.8 & 14.5 & 3.2 \\
\bottomrule
\end{tabular}
\end{table}

Three properties of the data shape every later choice. First, answers repeat heavily: the 134,416 training rows contain only 65,649 distinct answer texts, and 64.5\% of rows share their answer with at least one other row. This is why a naive in-batch loss would push a question away from a perfectly good answer, and why the loss in Section~\ref{sec:method} masks rows with equal answer identifiers. It also means that a model can score above chance without reading the question, by preferring answers that were often correct in training; we measure this with a question-blind baseline.

Second, validation is easier than test. Validation is drawn from the training distribution, so its answers recur in training, while the test set is the official development set. The two differ in subject mix: dental questions are 29.4\% of the test set and only 5.1\% of the training set. We therefore report a macro average over the 21 subjects next to overall accuracy, and we never use test data to choose a model.

Third, about a fifth of the questions are negated (19.8\% to 21.4\%). A question that asks which option is \emph{not} true is hard for any method that rewards similarity, which we examine in Section~\ref{sec:results}.

\section{Experimental Setup and Methodology}
\label{sec:method}

\subsection{Overview}

\begin{figure}[t]
\centering
\resizebox{\linewidth}{!}{%
\begin{tikzpicture}[
  box/.style={draw, rounded corners=2pt, align=center, font=\footnotesize, minimum height=9mm, minimum width=19mm, inner sep=3pt},
  vec/.style={draw, rounded corners=2pt, align=center, font=\footnotesize, minimum height=9mm, minimum width=12mm, fill=best},
  arr/.style={-{Stealth[length=2mm]}, thick}]
  \node[box] (q) {Question};
  \node[box, right=5mm of q] (qp) {Question\\prompt};
  \node[box, right=5mm of qp, minimum width=27mm] (lm1) {Frozen LM:\\last token at\\several depths};
  \node[box, right=5mm of lm1] (h1) {State\\head};
  \node[vec, right=5mm of h1] (s) {$\mathbf{s}$\\512-d};
  \node[box, below=13mm of q] (o) {Option};
  \node[box, right=5mm of o] (op) {Option\\prompt};
  \node[box, right=5mm of op, minimum width=27mm] (lm2) {Frozen LM:\\last token at\\several depths};
  \node[box, right=5mm of lm2] (h2) {Action\\head};
  \node[vec, right=5mm of h2] (a) {$\mathbf{a}_k$\\512-d};
  \node[box, fill=best, minimum width=26mm] (sc) at ($(s)!0.5!(a)+(3.4,0)$) {scale $\cdot\cos(\mathbf{s},\mathbf{a}_k)$\\softmax over options\\pick the largest};
  \foreach \x/\y in {q/qp, qp/lm1, lm1/h1, h1/s, o/op, op/lm2, lm2/h2, h2/a}
    \draw[arr] (\x) -- (\y);
  \draw[arr] (s.east) -- (sc.north west);
  \draw[arr] (a.east) -- (sc.south west);
\end{tikzpicture}}
\caption{The model. Only the two heads and one scale value are trained; the language model is frozen. All options of a question share one action head, and their vectors can be cached.}
\label{fig:pipeline}
\end{figure}
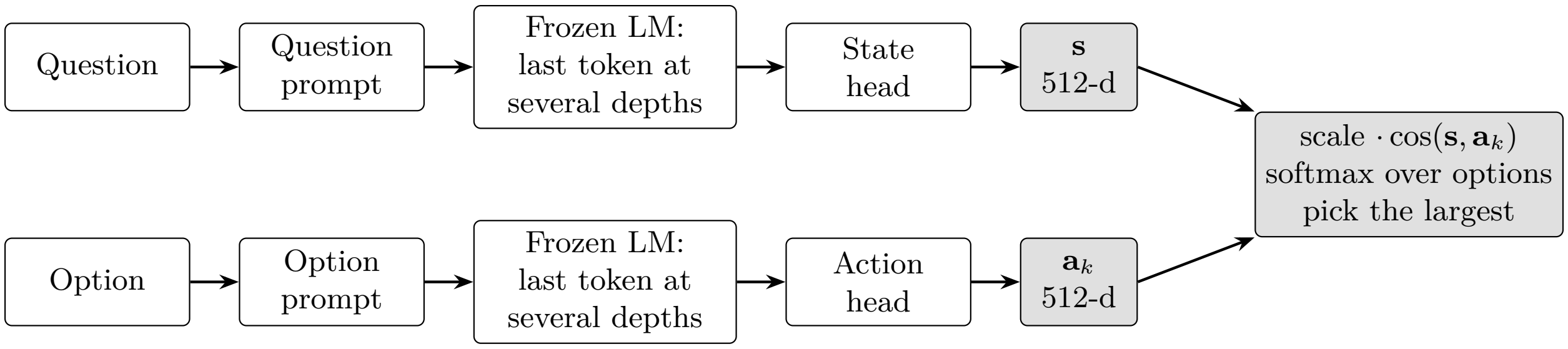

The pipeline has three stages (Figure~\ref{fig:pipeline}). A frozen encoder turns every unique question and option into a vector, once. Two small heads are then trained on the saved vectors, so training touches no language model and a run takes minutes. Finally, the trained heads score the options of each test question. We wrote all code ourselves, using the public CLM rebuild \citep{vizuara2026} only as a guide to the method.

\subsection{Frozen encoders}

We use three open models, all under Apache-2.0 (Table~\ref{tab:encoders}). Qwen3.5-0.8B \citep{qwen2026qwen35} is a causal language model with a vision encoder, of which we load only the language part; it alternates Gated DeltaNet and gated attention blocks. Granite-4.2-3B \citep{ibm2026granite} is a dense decoder-only transformer. Qwen3-8B \citep{yang2025qwen3} is the model family the CLM rebuild uses. Each model is loaded in bfloat16 through the Transformers library \citep{wolf2020transformers}, switched to evaluation mode, and never updated.

\begin{table}[t]
\centering
\caption{Frozen encoders and the trainable heads built on them. Vector size is for the prompted setting used in the final runs. Head weights count both heads and the scale value at hidden size 2048.}
\label{tab:encoders}
\begin{tabular}{lrrrrr}
\toprule
Encoder & Layers & Width & Parameters & Vector size & Head weights \\
\midrule
Qwen3.5-0.8B & 24 & 1,024 & 0.75B (language part) & 3,072 (3 depths)  & 23.1M \\
Granite-4.2-3B & 40 & 2,560 & 3.66B & 20,480 (8 depths) & 94.4M \\
Qwen3-8B & 36 & 4,096 & 8.2B & 12,288 (3 depths) & 60.8M \\
\bottomrule
\end{tabular}
\end{table}

\subsection{From a text to a vector}

A text is turned into a vector in three steps, and we vary the first and the third.

\paragraph{Input formulation.} In the \emph{plain} setting the text is fed as it is. In the \emph{eol} setting it is wrapped in a one-word-summary prompt in the style of PromptEOL \citep{jiang2024prompteol}: a question becomes \texttt{This question: "[text]" is answered in one word: "} and an option becomes \texttt{This answer: "[text]" means in one word: "}. In the \emph{qa} setting a question becomes \texttt{Question: [text]} followed by a line \texttt{Answer:}, and an option becomes \texttt{Answer: [text]}. The \emph{eol8} setting uses the eol prompts but reads more depths.

\paragraph{Pooling.} Texts are padded on the right and truncated at 512 tokens. We take the hidden state of the last non-padding token, found from the attention mask. This avoids relying on left padding, which behaves differently in recurrent layers.

\paragraph{Depth.} The plain setting reads the final layer only. The eol and qa settings read three depths, at 0.5, 0.75 and 1.0 of the model's depth, and eol8 reads eight depths (0.25, 0.375, 0.5, 0.625, 0.75, 0.875, 0.9375 and 1.0). For a model with $L$ layers, depth $f$ means the hidden state at index $\mathrm{round}(fL)$, where index 0 is the embedding output. Each depth's vector is L2-normalised and the vectors are concatenated. Vectors are computed in float32 and stored in float16.

Embedding covers 384,435 unique texts for plain and 384,591 once a template is applied. We computed them once per configuration and reused them for every head experiment.

\subsection{Heads and scoring}

The question goes through a state head and each option through an action head. A head is $\mathrm{Linear}(d,h)\rightarrow\mathrm{GELU}\rightarrow\mathrm{Dropout}\rightarrow\mathrm{Linear}(h,h)\rightarrow\mathrm{LayerNorm}\rightarrow\mathrm{GELU}\rightarrow\mathrm{Dropout}\rightarrow\mathrm{Linear}(h,512)$, using GELU \citep{hendrycks2016gelu} and layer normalisation \citep{ba2016layernorm}; dropout is zero unless stated. The 512-dimensional output is L2-normalised. The score of option $k$ is $\sigma\,\cos(\mathbf{s},\mathbf{a}_k)$, where $\sigma=\min(e^{\theta},100)$ and the learned parameter $\theta$ starts at $\ln(1/0.07)$. The initial recipe used hidden size $h=1024$; the final runs use $h=2048$.

\subsection{Loss}

For a batch of $B$ questions, let $\mathbf{s}_i$ be the state vector of question $i$, $\mathbf{a}_i$ the vector of its correct answer, $\mathbf{n}_{ik}$ its $k$-th wrong option ($k\le 3$), and $g_i$ the answer identifier. The question-to-answer logits are
\begin{equation}
\ell_{ij}=\sigma\,\mathbf{s}_i^{\top}\mathbf{a}_j \quad (j\neq i,\ g_j\neq g_i),\qquad \ell_{ii}=\sigma\,\mathbf{s}_i^{\top}\mathbf{a}_i,\qquad \ell^{\mathrm{hard}}_{ik}=\sigma\,\mathbf{s}_i^{\top}\mathbf{n}_{ik},
\end{equation}
where rows with the same answer identifier as $i$ are removed from its candidates, and padded negatives are ignored. Cross-entropy over $\{\ell_{ij}\}_j\cup\{\ell^{\mathrm{hard}}_{ik}\}_k$ with target $i$ gives $\mathcal{L}_{q\to a}$. The reverse direction ranks the questions for each answer, with the same mask and no hard negatives, and gives $\mathcal{L}_{a\to q}$. The loss is $\tfrac12(\mathcal{L}_{q\to a}+\mathcal{L}_{a\to q})$. The wrong options of a question are therefore used as extra candidates only in the direction where they make sense, and they are not shared with other questions. This follows the InfoNCE form \citep{oord2018cpc} with hard negatives in the spirit of \citet{karpukhin2020dpr}.

\subsection{Optimisation and model selection}

We train with AdamW \citep{loshchilov2019adamw}, a constant learning rate, batch size 4,096 (33 steps per epoch), gradient-norm clipping at 1.0 and weight decay 0 unless stated. The learning rate is 0.001 in the initial recipe and 0.0003 in the final one. Training stops when validation loss has not improved for five epochs, with at most 100 epochs, and the best-epoch weights are kept. A seed fixes the head initialisation and the shuffling; final results use seeds 42, 43 and 44.

Every choice (prompt, depth, learning rate, dropout, hidden size, batch size, weight decay) was made on validation accuracy of a single seed-42 run. The test set was evaluated first for the initial recipe with plain vectors, and later once for each final configuration. We chose which model to publish from validation accuracy and validation loss, and test accuracy agreed with that choice. The order matters for the reading of Section~\ref{sec:results}: the plain-vector test numbers use the initial recipe, which we had not yet tuned.

\subsection{Evaluation protocol}

Accuracy is the share of questions where the correct answer scores highest among that question's own options. If the best score is shared by several options, the correct option earns an equal share of the credit, so the result does not depend on where the correct option sits in a list. We report overall accuracy, the macro average over the 21 subjects, and accuracy on negated and non-negated questions. For a single run on 3,842 questions the 95\% margin is about $\pm1.6$ points; on 4,000 validation questions it is about $\pm1.5$.

Three reference points frame every result. \emph{Chance} is the mean of $1/(\text{number of options})$. The \emph{question-blind baseline} picks the option that was most often the correct answer in training and never reads the question. The \emph{no-head baseline} takes the cosine of the raw frozen vectors with no training. A fourth check, the \emph{shuffled-question control}, pairs each validation question with the options of another question and scores it; a model that reads the question should fall to chance.

\subsection{Implementation}

All runs used PyTorch 2.15.0 \citep{paszke2019pytorch}, Transformers 5.23.0, Datasets 5.2.0 and Python 3.14 on two NVIDIA L40 GPUs, one encoder per GPU. The two released models are public \citep{mantha2026qwen3model,mantha2026granitemodel}; the pipeline code is not yet released.

\section{Results and Discussion}
\label{sec:results}

\subsection{Test accuracy}

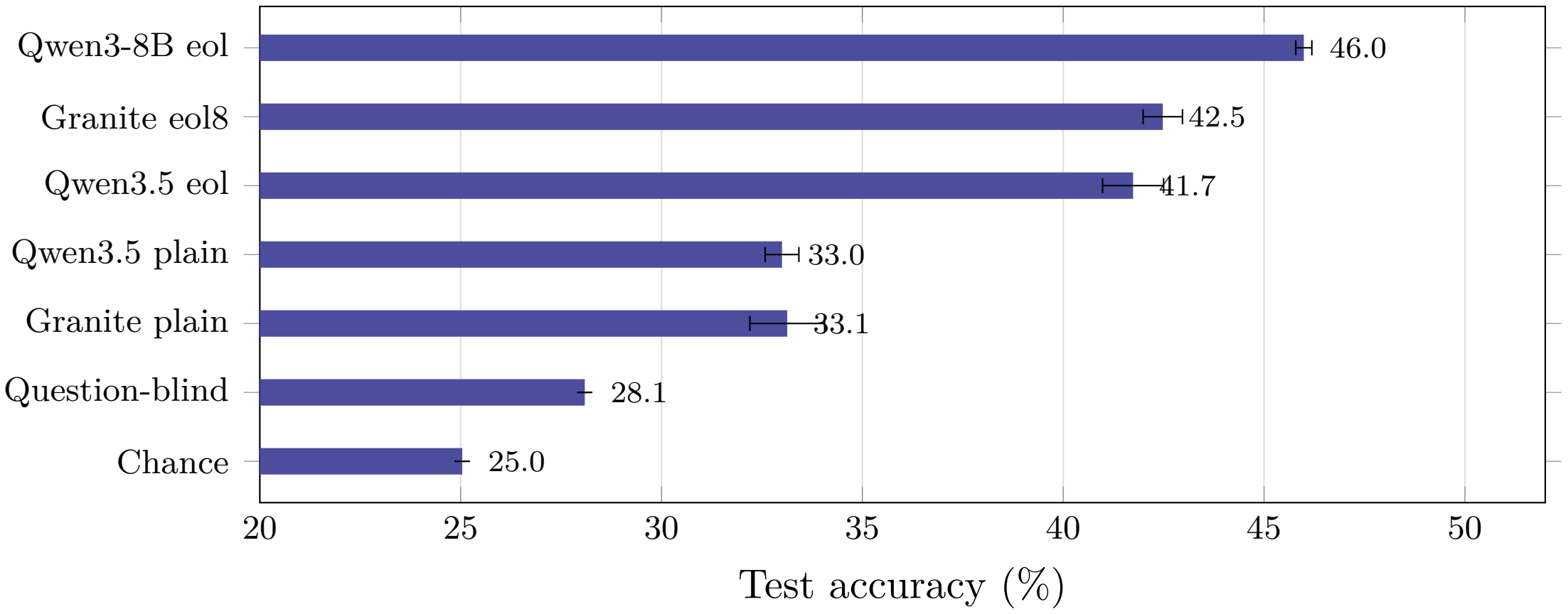
\begin{figure}[t]
\centering
\begin{tikzpicture}
\begin{axis}[
  xbar, width=0.82\linewidth, height=6.2cm,
  xmin=20, xmax=52, xlabel={Test accuracy (\%)},
  symbolic y coords={Qwen3-8B eol,Granite eol8,Qwen3.5 eol,Qwen3.5 plain,Granite plain,Question-blind,Chance},
  ytick=data, y dir=reverse, bar width=7pt, tick label style={font=\footnotesize},
  nodes near coords, nodes near coords style={font=\scriptsize, anchor=west, xshift=3pt},
  every node near coord/.append style={/pgf/number format/.cd, fixed, fixed zerofill, precision=1},
  enlarge y limits=0.1, xmajorgrids, major grid style={gray!25}]
\addplot[fill=blue!45!black!70, draw=none, error bars/.cd, x dir=both, x explicit]
  coordinates {(45.99,Qwen3-8B eol) +- (0.20,0) (42.48,Granite eol8) +- (0.49,0) (41.74,Qwen3.5 eol) +- (0.76,0)
               (33.00,Qwen3.5 plain) +- (0.42,0) (33.13,Granite plain) +- (0.93,0)
               (28.09,Question-blind) +- (0,0) (25.04,Chance) +- (0,0)};
\end{axis}
\end{tikzpicture}
\caption{Test accuracy by configuration. Error bars show the standard deviation over three seeds. ``Plain'' rows use the initial recipe.}
\label{fig:bars}
\end{figure}

Table~\ref{tab:main} and Figure~\ref{fig:bars} give the test results. With the initial recipe and plain last-token vectors, Granite-4.2-3B and Qwen3.5-0.8B are indistinguishable at 33.1\% $\pm$ 0.9 and 33.0\% $\pm$ 0.4, a few points above the question-blind baseline (28.1\%) and well above chance (25.0\%). With plain vectors, the raw frozen encoders with no training score close to chance (25.8\% to 27.0\%), so the heads are doing real work. Prompting also lifts the untrained cosine, to between 28.7\% and 34.0\% depending on the encoder; the untrained Qwen3-8B vectors with the eol prompt (34.0\%) already match the trained plain-vector models. With the final recipe, accuracy reaches 41.7\% for Qwen3.5-0.8B, 42.5\% for Granite-4.2-3B and 46.0\% for Qwen3-8B. The seed-to-seed standard deviation is 0.2 to 0.9 points, smaller than the single-run margin of about 1.6, so the gaps of three to four points between Qwen3-8B and the other two are larger than the noise we can measure.

\begin{table}[t]
\centering
\caption{Test accuracy (\%, $\uparrow$) on 3,842 questions. Trained rows show mean $\pm$ standard deviation over three seeds. ``Initial'' is the first recipe (hidden size 1,024, learning rate 0.001, plain vectors); ``final'' uses prompted vectors, hidden size 2,048 and learning rate 0.0003. Slices are averaged over seeds.}
\label{tab:main}
\small
\setlength{\tabcolsep}{4pt}
\begin{tabular}{llcccc}
\toprule
Method & Setting & Overall & Subject mean & EXCEPT/NOT & Other \\
\midrule
Chance                      & --                  & 25.04 & --    & --    & -- \\
Question-blind              & training frequency  & 28.09 & 28.86 & 28.26 & 28.04 \\
\addlinespace
No heads, Granite-4.2-3B    & plain               & 27.04 & --    & --    & -- \\
No heads, Qwen3.5-0.8B      & plain               & 25.77 & --    & --    & -- \\
No heads, Qwen3.5-0.8B      & eol                 & 28.71 & 28.84 & 24.18 & 29.83 \\
No heads, Granite-4.2-3B    & eol8                & 30.79 & --    & --    & -- \\
No heads, Qwen3-8B          & eol                 & 33.97 & --    & --    & -- \\
\addlinespace
Trained, Granite-4.2-3B     & plain, initial      & 33.13 $\pm$ 0.93 & 32.44 & 27.55 & 34.50 \\
Trained, Qwen3.5-0.8B       & plain, initial      & 33.00 $\pm$ 0.42 & 31.66 & 27.64 & 34.33 \\
Trained, Qwen3.5-0.8B       & eol, final          & 41.74 $\pm$ 0.76 & 44.22 & 37.49 & 42.79 \\
Trained, Granite-4.2-3B     & eol8, final         & 42.48 $\pm$ 0.49 & 45.90 & 35.79 & 44.13 \\
Trained, Qwen3-8B           & eol, final          & \cellcolor{best}45.99 $\pm$ 0.20 & \cellcolor{best}47.40 & 34.87 & \cellcolor{best}48.74 \\
\bottomrule
\end{tabular}
\end{table}

\subsection{What produced the gain}

The jump from 33\% to 42--46\% mixes several changes, so we separate them on validation data (Tables~\ref{tab:heads} and~\ref{tab:inputs}). Each cell is one seed-42 run, so differences below about two points should be treated as noise.

\paragraph{The learning rate of the heads.} On plain Granite vectors, moving from the initial learning rate of 0.001 to 0.0003 lifted validation accuracy from 37.5\% to 45.3\%, the largest single effect among the head settings we tried (Table~\ref{tab:heads}). On Qwen3.5-0.8B the same change gave 40.0\% to 43.7\%. Dropout of 0.3 also helped (to 42.9\% and 41.8\%). A smaller hidden size and weight decay gave gains of up to three points on Granite and almost none on Qwen3.5-0.8B, and a larger hidden size and smaller batches gave no clear gain.

\begin{table}[t]
\centering
\caption{Head settings on plain last-layer vectors, validation accuracy (\%, $\uparrow$, seed 42). Every row changes the initial recipe (learning rate 0.001, batch 4,096, hidden 1,024) in the way named.}
\label{tab:heads}
\begin{tabular}{lcc}
\toprule
Setting & Granite-4.2-3B & Qwen3.5-0.8B \\
\midrule
Initial recipe               & 37.45 & 40.02 \\
Learning rate 0.0003         & \cellcolor{best}45.32 & \cellcolor{best}43.73 \\
Dropout 0.3                  & 42.95 & 41.83 \\
Dropout 0.5                  & 40.47 & 39.27 \\
Weight decay 0.05            & 39.50 & 40.27 \\
Hidden size 512              & 40.45 & 40.27 \\
Hidden size 2,048            & 36.60 & 37.20 \\
Batch size 1,024             & 38.70 & 36.90 \\
Batch size 256               & 38.67 & 37.82 \\
Batch 1,024 and dropout 0.3  & 41.15 & 43.45 \\
Batch 256 and dropout 0.5    & 41.00 & 40.20 \\
\bottomrule
\end{tabular}
\end{table}

\paragraph{The input formulation.} Wrapping texts in a one-word-summary prompt helped more than any head setting. With the same single layer and the initial learning rate, Granite's validation accuracy rose from 37.5\% (plain) to 47.1\% (qa) and 50.1\% (eol), so the prompt alone is worth about 10 to 13 points (Table~\ref{tab:inputs}). With the learning rate matched at 0.0003, the plain setting reached 45.3\% and the eol setting with three depths reached 55.3\%, a gain of 10.0 points for Granite and 9.2 points for Qwen3.5-0.8B (43.7\% to 53.0\%). The shape of that comparison is worth stating plainly: it is the combined effect of the prompt and the extra depths, and we did not run the prompted setting with one depth at the lower rate, so we cannot say how much of the 10 points each part contributes.

\begin{table}[t]
\centering
\caption{Input formulation, depth and learning rate on validation accuracy (\%, $\uparrow$, seed 42). All rows use hidden size 1,024 except the last block. ``Depths'' are fractions of the encoder's depth. The rows differ in more than one factor, as the first three columns show.}
\label{tab:inputs}
\begin{tabular}{llccc}
\toprule
Format & Depths & Learning rate & Granite & Qwen3.5 \\
\midrule
plain & 1.0                    & 0.001  & 37.45 & 40.02 \\
plain & 1.0                    & 0.0003 & 45.32 & 43.73 \\
\addlinespace
qa    & 1.0                    & 0.001  & 47.10 & -- \\
qa    & 0.75, 1.0              & 0.001  & 46.45 & -- \\
qa    & 0.5                    & 0.001  & 37.00 & -- \\
qa    & 0.5, 0.75, 1.0         & 0.0003 & 53.53 & -- \\
\addlinespace
eol   & 1.0                    & 0.001  & 50.10 & -- \\
eol   & 0.75, 1.0              & 0.001  & 50.25 & -- \\
eol   & 0.5                    & 0.001  & 24.85 & -- \\
eol   & 0.5, 0.75, 1.0         & 0.0003 & \cellcolor{best}55.33 & \cellcolor{best}52.95 \\
eol8  & eight depths           & 0.0003 & 54.50 & -- \\
\midrule
\multicolumn{5}{l}{\emph{Final configuration (hidden size 2,048, learning rate 0.0003)}} \\
eol8  & eight depths           & 0.0003 & 56.05 & -- \\
eol   & 0.5, 0.75, 1.0         & 0.0003 & -- & 55.30 \\
\bottomrule
\end{tabular}
\end{table}

\paragraph{What we could not separate.} We expected that reading several depths would help, but our data do not show it. Adding the 0.75-depth state to the final layer changed nothing (50.1\% to 50.3\% for eol, 47.1\% to 46.5\% for qa), and going from three depths to eight did not help (55.3\% to 54.5\%). The only rows with three depths also use the lower learning rate, so the benefit of depth is confounded with the learning rate. The middle layer alone failed to train at the higher rate (24.9\% for eol, which is chance, with the best validation loss reached at epoch 2), but we did not repeat it at the lower rate, so we do not know whether the middle layer carries little information or the heads simply failed to fit it. A matched ablation, with the other factors fixed, is the first thing we would add.

\paragraph{Head settings on richer vectors.} Once the vectors improved, the best head settings changed (Table~\ref{tab:richer}). A hidden size of 2,048 now helped by 1.5 points for Granite and 2.4 for Qwen3.5-0.8B, where it had hurt on plain vectors. Dropout, which had helped on plain vectors, now hurt by about 5 points. Smaller batches hurt too. This is why we tuned the head again after changing the vectors, and it suggests that head settings found on one kind of vector should not be carried over to another.

\begin{table}[t]
\centering
\caption{Head settings on prompted vectors, validation accuracy (\%, $\uparrow$, seed 42). Rows use hidden size 1,024 and learning rate 0.0003 unless a different setting is named.}
\label{tab:richer}
\begin{tabular}{llc}
\toprule
Vectors & Setting & Accuracy \\
\midrule
Granite, eol (3 depths)  & learning rate 0.0003              & 55.33 \\
                         & learning rate 0.0003, dropout 0.3 & 49.95 \\
                         & learning rate 0.0001, dropout 0.3 & 51.05 \\
\addlinespace
Granite, eol8            & learning rate 0.0003              & 54.50 \\
                         & learning rate 0.0001              & 55.13 \\
                         & batch size 1,024                  & 51.40 \\
                         & weight decay 0.05                 & 55.43 \\
                         & hidden size 2,048                 & \cellcolor{best}56.05 \\
\addlinespace
Qwen3.5-0.8B, eol        & learning rate 0.0003              & 52.95 \\
                         & learning rate 0.0001              & 53.23 \\
                         & batch size 1,024                  & 50.62 \\
                         & weight decay 0.05                 & 51.88 \\
                         & hidden size 2,048                 & \cellcolor{best}55.30 \\
\bottomrule
\end{tabular}
\end{table}

\paragraph{How much of the gain is real.} Validation and test move together, but not equally. Between the initial recipe and the final one, Granite's validation accuracy rose from 37.5\% to 56.1\% (18.6 points) while its test accuracy rose from 33.1\% to 42.5\% (9.4 points). For Qwen3.5-0.8B the figures are 15.3 points on validation and 8.7 on test. Validation questions share their answers with the training set, and test questions come from a different distribution, so validation overstates how well the heads transfer. Anyone using validation accuracy to rank configurations should expect the ordering to carry over better than the size of the gap.

\subsection{Does the model use the question?}

A similarity model can score above chance for the wrong reason, so we ran the shuffled-question control on every validation run in our sweeps (44 runs). Accuracy fell to between 22.7\% and 26.7\%, with a mean of 24.2\%, against 24.9\% to 56.1\% on the same runs unshuffled (the 24.9\% run is the middle-layer model that failed to train). The heads therefore read the question and are not only exploiting how often an answer is correct. The question-blind baseline is a stricter check: it reaches 28.1\% on the test set, and our trained models beat it by 5 points (initial recipe) to 18 points (Qwen3-8B).

\subsection{Encoder choice}

With plain vectors and the initial recipe, a 0.8B and a 3.7B encoder tie (33.0\% against 33.1\%), so more parameters did not help there. With prompted vectors and the final recipe the three encoders separate: 41.7\% for Qwen3.5-0.8B, 42.5\% for Granite-4.2-3B and 46.0\% for Qwen3-8B, the best on both validation (57.3\%) and test. The first two differ by 0.7 points, about one standard deviation of the seed noise, while Qwen3-8B sits more than three points above both. We cannot attribute this to scale alone, because the encoders also differ in family, training data, vector size (3,072, 20,480 and 12,288) and number of depths. All we can say is that the largest model is the best of the three, and that the benefit of size appeared only after the vectors were improved.

\subsection{Where the model fails}

Questions phrased with \emph{except}, \emph{not}, \emph{false}, \emph{incorrect} or \emph{least} are much harder: Qwen3-8B reaches 34.9\% on them and 48.7\% on the rest, and a gap of 5 to 14 points appears for every trained model. We did not test the cause. A plausible one is that a question about what is \emph{not} true is embedded close to the topic, which makes the option that \emph{is} true look most similar, but we have not checked it.

Accuracy also varies by subject. For Qwen3-8B, averaged over seeds, it is highest for Pharmacology (59.7\%, $n=231$), Biochemistry (56.7\%, $n=150$) and Pathology (54.4\%, $n=321$), with Psychiatry higher still at 60.4\% but on only 16 questions, and lowest for Dental (37.6\%, $n=1{,}130$), Anaesthesia (34.3\%, $n=34$) and Orthopaedics (26.7\%, $n=20$); the last two groups are too small to read much into. Dental is 29.4\% of the test set, so it pulls the overall accuracy down: the macro average over subjects is 47.4\%, against 46.0\% overall, and the non-dental questions alone score about 49.5\%.

\subsection{Cost}

The pipeline is cheap by design (Table~\ref{tab:cost}). Embedding is the only stage that touches a language model, and it ran at 288 texts per second for Qwen3-8B and 780 to 1,390 for Granite. Training the heads on saved vectors took under three minutes per seed except for the 20,480-dimensional Granite vectors. We did not run a latency benchmark against generative models. A single question took about 31 seconds with the published Granite model on a 16~GB laptop CPU in bfloat16, so the practical speed advantage of a single-pass design depends on caching option vectors and on a GPU, which we did not measure here.

\begin{table}[t]
\centering
\caption{Wall-clock cost on two NVIDIA L40 GPUs. Embedding covers all 384,435 or 384,591 unique texts; training is per seed, to early stopping.}
\label{tab:cost}
\begin{tabular}{lrr}
\toprule
Configuration & Embedding (min) & Head training per seed (min) \\
\midrule
Granite-4.2-3B, plain & 8.4  & 0.4 to 0.5 \\
Qwen3.5-0.8B, plain   & 8.0  & about 0.4 \\
Qwen3.5-0.8B, eol     & 9.2  & 1.9 to 2.0 \\
Granite-4.2-3B, eol8  & 13.3 & 5.7 to 6.8 \\
Qwen3-8B, eol         & 28.7 & 2.4 to 2.9 \\
\bottomrule
\end{tabular}
\end{table}

\subsection{Comparison with published results}

Published numbers are far above ours: Flan-PaLM 540B reports 57.6\% on the MedMCQA development set \citep{singhal2022arxiv} and Med-PaLM~2 reports 72.3\% \citep{singhal2023arxiv}. These are not like-for-like. Ours is a four-way choice among a question's own options on 3,842 filtered questions, with no generation and with an encoder one to three orders of magnitude smaller, while the published figures come from large generative models under their own prompting protocols. We read the gap as the price of a frozen small encoder that never sees a question and its options together, not as a result that competes with those systems.

\subsection{Limitations}

Several limits should be kept in mind. The ablations are single-seed on 4,000 validation questions, with a margin of about 1.5 points, so small differences in Tables~\ref{tab:heads}--\ref{tab:richer} are not evidence. The test accuracy of the plain-vector models comes from the untuned initial recipe, and we did not evaluate plain vectors with the final head settings on the test set, so the test gain from better vectors alone is smaller than the 9 to 13 points the table suggests, by some amount we did not measure. The effect of reading several depths is not isolated. The test set is the MedMCQA development set, and since it is public we cannot rule out that the encoders saw it during pretraining. We used one dataset in English and three encoders from two families. The processed dataset removes questions with context-dependent options, which makes our task somewhat easier than the full benchmark. Finally, we lost access to the GPU machine before we could rerun the missing matched ablations.

\section{Future Scope of Work}
\label{sec:future}

The most direct next step is to let the encoder learn a little. Low-rank adaptation \citep{hu2021lora} would train a small number of extra weights inside the language model with the same contrastive loss. We left the encoder frozen on purpose, to see how far the readout alone goes, and the gap to the published results suggests that is where the headroom is. Whether it closes that gap is open, and we would not expect it to reach the strongest generative systems.

Before that, the evidence in Section~\ref{sec:results} needs repairing. We would run the matched ablation that separates the prompt, the depth and the learning rate, report every ablation over several seeds with paired comparisons, and evaluate plain and prompted vectors on the test set under the same head settings. A learned weighting over depths, in place of fixed fractions, would turn the depth question into something the data can answer.

Two modelling ideas follow from the failures we saw. Negated questions could be handled explicitly, for instance by detecting them and flipping the ranking, or by training on rewritten pairs. And since the question and options are encoded apart, a light interaction between them, such as scoring the question together with one option at a time, might help; it would give up some of the caching that makes the single-pass design cheap, so the trade-off needs measuring.

Finally, the design should be tested elsewhere: on MedQA \citep{jin2020medqa} and on other languages and domains, against generative baselines under the same protocol, and with a proper latency and cost comparison that uses cached option vectors. A check for pretraining contamination on the MedMCQA development set would strengthen every number in this paper.

\bibliographystyle{unsrtnat}
\bibliography{references}

\end{document}